\documentclass[twoside,11pt]{article}
\usepackage{pgm}

\usepackage[utf8]{inputenc}
\inputencoding{utf8}

\usepackage{hyperref,color,soul,booktabs}
\setulcolor{blue}
\newcommand{\BibTeX}{\textsc{B\kern-0.1emi\kern-0.017emb}\kern-0.15em\TeX}
\usepackage{bm}
\usepackage{amsmath}
\usepackage{amssymb}
\usepackage{algorithmic}
\usepackage{booktabs,subcaption,amsfonts,dcolumn}
\newcolumntype{d}[1]{D..{#1}}
\ShortHeadings{Efficient Sensitivity Analysis in Bayesian Networks}{Ballester-Ripoll and Leonelli}

\begin{document}

% Title and authors
\title{You Only Derive Once (YODO): Automatic Differentiation for Efficient Sensitivity Analysis in Bayesian Networks}
\author{\Name{Rafael Ballester-Ripoll} \Email{rafael.ballester@ie.edu}\and
   \Name{Manuele Leonelli} \Email{manuele.leonelli@ie.edu}\\
   \addr School of Science and Technology, IE University, Madrid, Spain}

\maketitle

% Abstract and keywords
\begin{abstract}%   <- trailing '%' for backward compatibility of .sty file
Sensitivity analysis measures the influence of a Bayesian network's parameters on a quantity of interest defined by the network, such as the probability of a variable taking a specific value. In particular, the so-called sensitivity value measures the quantity of interest's partial derivative with respect to the network's conditional probabilities. However, finding such values in large networks with thousands of parameters can become computationally very expensive. We propose to use automatic differentiation combined with exact inference to obtain all sensitivity values in a single pass. Our method first marginalizes the whole network once using e.g. variable elimination and then backpropagates this operation to obtain the gradient with respect to all input parameters. We demonstrate our routines by ranking all parameters by importance on a Bayesian network modeling humanitarian crises and disasters, and then show the method's efficiency by scaling it to huge networks with up to 100'000 parameters. An implementation of the methods using the popular machine learning library PyTorch is freely available.
\end{abstract}

\section{Introduction}

Probabilistic graphical models, and specifically Bayesian networks (BNs), are a class of models that are widely used for risk assessment of complex operational systems in a variety of domains. The main reason for their success is that they provide an efficient as well as intuitive framework to represent the joint probability of a vector of variables of interest using a simple graph.  Their use to assess the reliability of engineering, medical and ecological systems, among many others,  is becoming increasingly popular.  Sensitivity analysis is a critical step for any applied real-world analysis to assess the importance of various risk factors and to evaluate the overall safety of the system under study \citep[see e.g.][for some recent examples]{goerlandt2021bayesian,makaba2021bayesian,zio2022bayesian}. 
	
As noticed by  \citet{rohmer2020uncertainties}, sensitivity analysis in BNs is usually \textit{local}, in the sense that it measures the effect of a small number of parameter variations on output probabilities of interest, while other parameters are kept fixed. In the case of a single parameter variation, sensitivity analysis is usually referred to as \textit{one-way}, otherwise, when more than one parameter is varied, it is called \textit{multi-way}. Although recently there has been an increasing interest in proposing \textit{global} sensitivity methods for BNs measuring how different factors \textit{jointly} influence some function of the model's output \citep[see e.g.][]{ballester2022computing,li2018sensitivity}, the focus of this paper still lies in one-way sensitivity methods. However, extensions to multi-way local methods are readily available and discussed in Section \ref{sec:discussion}.

One-way local sensitivity analysis in BNs can be broken down into two main steps. First, some parameters of the model are varied and the effect of these variations on output probabilities of interest is investigated. For this purpose, a simple mathematical function, usually termed \emph{sensitivity function}, describes an output probability of interest as a function of the BN parameters \citep{castillo1997sensitivity,coupe2002properties}. Furthermore, some specific properties of such a function can be computed, as for instance, the \emph{sensitivity value} or the \emph{vertex proximity}, which give an overview of how sensitive the probability of interest is to variations of the associated parameter. Such measures are reviewed below in Section \ref{sec:sens}. Second, once parameter variations are identified, the effect of these is summarized by a distance or divergence measure between the original and the varied distributions underlying the BN, most commonly the Chan-Darwiche distance \citep{chan2005distance} or the well-known Kullback-Leibler divergence. 
	
As demonstrated by \citet{uai2008}, the derivation of both the sensitivity function and its associated properties is computationally very demanding. Here we provide a novel, computationally highly-efficient method to compute all sensitivity measures of interest which takes advantage of backpropagation and is easy to compute thanks to automatic differentiation. Our algorithm is demonstrated in a BN modeling humanitarian crises and disasters (Sec.~\ref{sec:appl}), and an extensive simulation study shows its efficiency by processing huge networks in a few seconds. We have open-sourced a Python implementation using the popular machine learning library PyTorch \footnote{Available at \url{https://github.com/rballester/yodo}.}, contributing to the recent effort of promoting sensitivity analysis \citep{Douglas2020}.

\section{Bayesian Networks and Sensitivity Analysis}
\label{sec:sens}

A BN is a probabilistic graphical model defining a factorization of the probability mass function of a random vector by means of a directed acyclic graph (DAG). More formally, let $[n]=\{1,\dots,n\}$ and $\bm{Y}=(Y_i)_{i\in[n]}$ be a random vector of interest with sample space $\mathbb{Y}=\times_{i\in[n]}\mathbb{Y}_i$. A BN defines the probability mass function  $P(\bm{Y}=\bm{y})$, for $\bm{y}\in\mathbb{Y}$, as a product of simpler conditional probability mass functions as follows:
\begin{equation}
P(\bm{Y}=\bm{y}) = \prod_{i\in[n]}P(Y_i=y_i\;|\; \bm{Y}_{\Pi_i}=\bm{y}_{\Pi_i}),
\end{equation}
where $\bm{Y}_{\Pi_i}$ are the parents of $Y_i$ in the DAG associated to the BN.

The definition of the probability mass function over $\bm{Y}$, which would require defining $\#\mathbb{Y}-1$ probabilities, is thus simplified in terms of one-dimensional conditional probability mass functions. The coefficients of these functions are henceforth referred to as the parameters $\bm\theta$ of the model. The DAG structure may be either expert-elicited or learned from data using structural learning algorithms, and the associated parameters $\bm\theta$ can be either expert-elicited as well or learned using frequentist or Bayesian approaches.  No matter the method used, we assume that a value for these parameters $\bm\theta$ has been chosen which we refer to as the \textit{original value} and denoted as $\bm\theta^0$.

In practical applications it is fundamental to extensively assess the implications of the chosen parameter values $\bm\theta^0$ to outputs of the model. In the context of BNs this study is usually referred to as \textit{sensitivity analysis}, which can actually be further used during the model building process as showcased by \citet{coupe2000sensitivity}. Let $Y_O$ be an output variable of interest and $\bm{Y}_E$ be \textit{evidential} variables, those that may be observed. The interest is in then studying how $P(Y_O=y_O\;|\; \bm{Y}_E=\bm{y}_E)$ varies when a parameter $\theta_i$ is varied. In particular, $P(Y_O=y_O\;|\; \bm{Y}_E=\bm{y}_E)$ seen as a function of $\theta_i$ is called \textit{sensitivity function} and denoted as $f(\theta_i)$.

\subsection{Proportional Covariation}

Notice that when an input $\theta_i$ is varied from its original value $\theta_i^0$ then the parameters from the same conditional probability mass function need to \textit{covary} to respect the sum-to-one condition of probabilities. When variables are binary, this is automatic since one parameter must be equal to one minus the other, but for variables taking more than two levels this covariation can be done in several ways \citep{renooij2014co}. We henceforth assume that whenever a parameter is varied from its original value $\theta_i^0$  to a new value $\theta_i$, then every parameter $\theta_j$ from the same conditional probability mass function is \textit{proportionally covaried} \citep{laskey1995sensitivity} from its original value $\theta_j^0$:

\begin{equation} \label{eq:proportional_covariation}
	\theta_j(\theta_i) = \frac{1-\theta_i}{1-\theta^0_i}\theta^0_j.
\end{equation}

Proportional covariation has been studied extensively and its choice is motivated by a wide array of theoretical properties \citep{chan2005distance,leonelli2017sensitivity,leonelli2018geometric}.

Under the assumption of proportional covariation, \citet{castillo1997sensitivity} and \citet{coupe2002properties} demonstrated that the sensitivity function is the ratio of two linear functions:
\begin{equation} \label{eq:sens}
	f(\theta_i)=\frac{c_1\theta_i+c_2}{c_3\theta_i+c_4},
\end{equation}
where $c_1,c_2,c_3,c_4\in\mathbb{R}_{+}$. \citet{van2007sensitivity}  noticed that the above expression actually coincides with the fragment of a rectangular hyperbola, which can be generally written as 
\begin{equation}
	f(\theta_i) = \frac{r}{\theta_i-s}+t,
\end{equation}
where
	$s = -\frac{c_4}{c_3}$, $t=\frac{c_1}{c_3}$ and  $r = \frac{c_2}{c_3}+s t$.

\subsection{Sensitivity Value}
\label{sec:sensitivity_value}

The \textit{sensitivity value}  describes the effect of infinitesimally small shifts in the parameter’s original value on the probability of interest and is defined as the absolute value of the first derivative of the sensitivity function at the original value of the parameter, i.e. $|f^{'}(\theta_i^0)|$. This can be found by simply differentiating the sensitivity function as
\begin{equation} \label{eq:first_derivative}
|f^{'}(\theta_i^0)| = \frac{|c_1 c_4 - c_2 c_3|}{(c_3\theta_i^0+c_4)^2}.
\end{equation}
The higher the sensitivity value, the more sensitive the output probability to small changes in the original value of the parameter. As a rule of thumb, parameters having a sensitivity value larger than one may require further investigation.

Notice that  when $\bm{Y}_E$ is empty, i.e. the output probability of interest is a marginal probability, then the sensitivity function is linear in $\theta_i$ and the sensitivity value is the same no matter what the original $\theta_i^0$ was. Therefore in this case the absolute value of the gradient is sufficient to quantify the effect of a parameter to an output probability of interest.

\subsection{Vertex Proximity}
\label{sec:vertex_proximity}

\citet{van2007sensitivity} further noticed that parameters for which the sensitivity value is small may still be such that the conditional output probability of interest is very sensitive to their variations.  This happens when the original parameter value is close to the \textit{vertex} of the sensitivity function, defined as the point $\theta_i^v$ at which the sensitivity value is equal to one, i.e.
\begin{equation}
	|f^{'}(\theta_i^v)| = 1.
\end{equation}
The vertex can be derived from the equation of the sensitivity function as
\begin{equation}
	\theta_i^v = \left\{
	\begin{array}{ll}
	s+\sqrt{|r|}, & \mbox{if } s <0, \\
	s - \sqrt{|r|}, & \mbox{if } s > 0.
	\end{array}
	\right.
\end{equation}
Notice that the case $s=0$ is not contemplated since it would coincide to a linear sensitivity function, not an hyperbolic one.

\textit{Vertex proximity} is defined as the absolute difference $|\theta_i^0-\theta_i^v|$. The smaller the vertex proximity, the more sensitive the output probabilities may be to variations of the parameter, even when the sensitivity value is small. 

\subsection{Other Metrics}
\label{sec:other_metrics}

Given the coefficients $c_1, \dots, c_4$ of Eq.~\ref{eq:sens}, it is straightforward to derive any property of the sensitivity function besides the sensitivity value and the vertex proximity. Here we propose the use of two additional metrics.  The first is the absolute value of the second derivative of the sensitivity function at the original parameter value, which can be easily computed as: 
\begin{equation} \label{eq:second_derivative}
	|f''(\theta_i^0)| = \frac{2 c_{3} \left|c_{1} c_{4} - c_{2} c_{3}\right|}{\left(c_{3} \theta_i^0 + c_{4}\right)^{3}}.
\end{equation}
Similarly to the sensitivity value, high values of the second derivative at $\theta_i^0$ indicate parameters that could highly impact the probability of interest. 

The second measure  is the maximum of the first derivative of the sensitivity function over the interval $[0, 1]$ in absolute value, which we find easily by noting that the denominator of Equation~\ref{eq:first_derivative} is a parabola:
\begin{equation}
	\max_{\theta_i \in [0, 1]}\{|f'(\theta_i)|\} = \begin{cases}
		\infty & \mbox{ if } -c_4 / c_3 \in [0, 1] \\
		\max \{|c_1 c_4 - c_2 c_3| / c_4^2, |c_1 c_4 - c_2 c_3| / (c_3 + c_4)^2\} & \mbox{ otherwise.} \\
	\end{cases}
\end{equation}
Again high values indicate parameters whose variations can lead to a significant change in the output probability of interest.

\section{The YODO Method}

%Our method supports two types of sensitivity functions:
%
%\begin{itemize}
%	\item Marginal probability: $P(Y_O=y_O)$
%	\item Conditional probability: $P(Y_O=y_O\;|\; \bm{Y}_E=\bm{y}_E)$
%\end{itemize}
%
%We detail the algorithm for the second case, since the first one is a particular case by setting $\bm{Y}_E = \emptyset$.

\subsection{First Case: Marginal Probability as a Function of Interest}
\label{sec:marginal_case}

Suppose $f(\theta_i) = P(Y_O = y_O) = c_1 \theta_i + c_2$ assuming proportional covariation as $\theta_i$ moves. Let $\theta_{j_1}, \dots, \theta_{j_n}$ be the other parameters of the same conditional PMF as $\theta_i$, i.e. they are all bound by the sum-to-one constraint $\theta_i + \theta_{j_1} + \dots + \theta_{j_n} = 1$.
%Let $g(\theta_i)$ be $P(Y_O = y_O)$ \emph{without} the assumption of proportional covariation, i.e. $\theta_i$ moves alone.
First, we rewrite $f$ as
\begin{equation}
	f(\theta_i) = g(\theta_i, \theta_{j_1}(\theta_i), \dots, \theta_{j_n}(\theta_i))
\end{equation}
and we will show how to obtain $f'(\theta_i)$ provided that we can compute the gradient $\nabla g$ with respect to symbols $\theta_i, \theta_{j_1}, \dots, \theta_{j_n}$ (Sec.~\ref{sec:computing_the_gradient} for details on the latter).

By the generalized chain rule, it holds that
\begin{equation} \label{eq:generalized_chain_rule}
	f'(\theta_i) = \frac{\partial g}{\partial \theta_i} \cdot 1 + \frac{\partial g}{\partial \theta_{j_1}} \cdot \frac{d\theta_{j_1}}{d\theta_i} + \dots + \frac{\partial g}{\partial \theta_{j_n}} \cdot \frac{d \theta_{j_n}}{d\theta_i}.
%	\frac{df(\theta_i)}{d\theta_i} = \frac{dg(\theta_i)}{d\theta_i} + \frac{dg(\theta_{j_1}(\theta_i))}{d\theta_i} + \dots + \frac{dg(\theta_{j_n}(\theta_i))}{d\theta_i}.
\end{equation}

By deriving Eq.~\ref{eq:proportional_covariation}, we have that for all $1 \le m \le n$:

\begin{equation} \label{eq:derivative}
	\frac{d\theta_{j_m}}{d\theta_i} = \frac{-\theta_{j_m}^0}{1 - \theta_i^0}
\end{equation}

%Next, by applying the chain rule on Eq.~\ref{eq:proportional_covariation}, for all $1 \le m \le n$ we have
%
%\begin{equation} \label{eq:chain_rule}
%	\frac{dg(\theta_{j_m}(\theta_i))}{d\theta_i} = \frac{-\theta_{j_m}^0 \cdot \frac{dg(\theta_{j_m})}{d\theta_{j_m}}}{1-\theta_i^0}.
%\end{equation}
%
%Suppose we can compute the gradient $\nabla g = (\partial g / \partial \theta)_{\theta \in \bm{\theta}}$ (Sec.~\ref{sec:computing_the_gradient} details how). From Eqs.~\ref{eq:additive} and~\ref{eq:chain_rule} we find $f'(\theta_i)$ as follows:

%Putting Eqs.~\ref{eq:generalized_chain_rule} and~\ref{eq:derivative} together, we have
and, therefore,
\begin{equation}
	f'(\theta_i) = \frac{\partial g}{\partial \theta_i} - \frac{(\partial g / \partial \theta_{j_1}) \cdot \theta_{j_1}^0 + \dots + (\partial g / \partial \theta_{j_n}) \cdot \theta_{j_n}^0}{1 - \theta_i^0}.
\end{equation}

Last, since $f(\theta_i) = P(\bm{Y}_O = \bm{y}_O) = c_1 \theta_i + c_2$, we easily find the parameters $c_1, c_2$:

\begin{equation}
	\begin{cases}
		c_1 = f'(\theta_i^0) \\
		c_2 = P(\bm{Y}_O = \bm{y}_O) - c_1 \theta_i^0.
	\end{cases}
\end{equation}

\subsection{Second Case: Conditional Probability as a Function of Interest}

When $f(\theta_i) = P(Y_O=y_O\;|\; \bm{Y}_E=\bm{y}_E) = P(Y_O=y_O, \bm{Y}_E=\bm{y}_E) / P(\bm{Y}_E = \bm{y}_E)$, we simply repeat the procedure from Sec.~\ref{sec:marginal_case} twice:

\begin{enumerate}
	\item We first apply it to $P(Y_O=y_O, \bm{Y}_E=\bm{y}_E)$ to obtain $c_1$ and $c_2$;
	\item we then apply it to $P(\bm{Y}_E=\bm{y}_E)$ to obtain $c_3$ and $c_4$.
\end{enumerate}

\subsection{Computing the Gradient $\nabla g$}
\label{sec:computing_the_gradient}

%Note that $g(\theta_i, \theta_{j_1})

Let $\bm{Y}_K = \bm{y}_K$ be a subset of the network variables taking some evidence values (this could be $K = O$ or $K = O \cup E$, hence we cover the two cases above).

We start by moralizing the BN into a Markov random field (MRF) $\mathcal{M}$. This marries all variable parents together and, for each conditional probability table (now called \emph{potential}), drops the sum-to-one constraint; see e.g.~\citep{Darwiche2009} for more details. Next, we impose the evidence $\bm{Y}_K = \bm{y}_K$ by defining $\mathcal{M}^{\bm{Y}_K = \bm{y}_K}$ as a new MRF that results from substituting each potential $\Phi_{i_1, \dots, i_M}(x_{i_1}, \dots, x_{i_M})$ by a new potential $\widehat{\Phi}_{i_1, \dots, i_M}$ defined as follows:

\begin{equation}
\begin{split}\widehat{\Phi}_{i_1, \dots, i_M}(Y_{i_1} = x_{i_1}, \dots, Y_{i_M} = x_{i_M}) = \\
\begin{cases}
0 & \mbox{if } \exists m, k \; \vert \; i_m = k \wedge x_{i_m} \ne y_{i_m} \\
\Phi_{i_1, \dots, i_M}(Y_{i_1} = x_{i_1}, \dots, Y_{i_M} = x_{i_M}) & \mbox{otherwise} \\
\end{cases}
\end{split}
\end{equation}

In other words, we copy the original potential but zero-out all entries that do not honor the assignment of values $\bm{Y}_K = \bm{y}_K$. See Tab.~\ref{tab:potential_example} for an example using a bivariate potential.
\begin{table}[h]
	\begin{subtable}[h]{0.48\textwidth}
		\begin{tabular}{c | c | c | c} 
			& $Y_2 = 1$ & $Y_2 = 2$ & $Y_2 = 3$ \\ [0.5ex] 
			\hline
			$Y_1 = 1$ & 0.8 & 0.1 & 0.1 \\ 
			\hline
			$Y_1 = 2$ & 0.3 & 0.5 & 0.2 \\
			\hline
			$Y_1 = 3$ & 0.1 & 0.2 & 0.7 \\
			\hline
		\end{tabular}
		\caption{$\Phi_{1, 2}(y_1, y_2)$}
	\end{subtable}
	\begin{subtable}[h]{0.48\textwidth}
		\begin{tabular}{c | c | c | c} 
			& $Y_2 = 1$ & $Y_2 = 2$ & $Y_2 = 3$ \\ [0.5ex] 
			\hline
			$Y_1 = 1$ & 0 & 0 & 0.1 \\ 
			\hline
			$Y_1 = 2$ & 0 & 0 & 0.2 \\
			\hline
			$Y_1 = 3$ & 0 & 0 & 0.7 \\
			\hline
		\end{tabular}
		\caption{$\widehat{\Phi}_{1, 2}(y_1, y_2)$}
	\end{subtable}
	\caption{Left: example potential of an MRF $\mathcal{M}$ for variables $Y_1$ and $Y_2$, each with three levels $\{1, 2, 3\}$. Right: corresponding potential for $\mathcal{M}^{Y_2 = 3}$.}
	\label{tab:potential_example}
\end{table}

Intuitively, the modified MRF $\mathcal{M}^{\bm{Y}_K = \bm{y}_K}$ represents the unnormalized probability for all variable assignments that are compatible with $\bm{Y}_K = \bm{y}_K$. In particular, if $\mathcal{M}_{\bm{Y}_K}$ denotes the marginalization of a network $\mathcal{M}$ over all variables in $\bm{Y}_K$, we have that $(\mathcal{M}^{\bm{Y}_K = \bm{y}_K})_{\bm{Y}} = P(\bm{Y}_K = \bm{y}_K)$. In other words, computing $g$ reduces to marginalizing our MRF. In this paper we marginalize it exactly using the variable elimination (VE) algorithm; see e.g.~\citep{Darwiche2009}. This method is clearly differentiable w.r.t. all parameters $\bm{\theta}$ since VE only relies on variable summation and factor multiplication. Any other differentiable inference algorithm could be used as well. This step, evaluating the function $g$, is known as the \emph{forward pass} in the neural network literature. Next, we backpropagate the previous operation (a step also known as the \emph{backward pass}) to build the gradient $\nabla g$. Crucially, note that backpropagation yields $\partial g / \partial \theta$ for every parameter $\theta \in \bm{\theta}$ of the network at once, not just an individual $\theta_i$. Last, we obtain parameters $c_1, \dots, c_4$ as detailed before, and use them to compute the metrics of Secs.~\ref{sec:sensitivity_value},~\ref{sec:vertex_proximity}, and~\ref{sec:other_metrics} for each $\theta_i$.

Note the advantages of this approach as compared to other alternatives. For example, symbolically deriving the gradient of $g$ would be cumbersome and would depend on the target network topology and definition of the probability of interest \citep{darwiche2003differential}. Automatic differentiation avoids this by evaluating the gradient numerically using the chain rule. Furthermore, finding the gradient using finite differences would require evaluating $g$ twice per parameter $\theta_i$. In contrast, automatic differentiation only requires a forward and backward pass to find the entire gradient --in our experiments, this took roughly the time of just two marginalization operations (see next section).

%\item Evaluate $f$ using an inference algorithm (evaluating the function is known as the \emph{forward pass} in the deep learning literature). In this paper we query $f$ via exact inference using variable elimination, which clearly yields a differentiable result since it only relies on variable summation and factor multiplication. Note that any other differentiable algorithm could be used as well.

\section{Results}
\label{sec:appl}
We overview first the insights revealed by our method when applied on a 21-node Bayesian network of interest; we then study the method's scalability by testing it on large networks with hundreds of nodes and arcs and up to $10^5$ parameters.

\subsection{Software and Hardware Used}

In order to perform variable elimination efficiently, we note that the problem of graphical model marginalization is equivalent to that of tensor network contraction~\cite{RS:18}, and use the library \emph{opt\_einsum}~\cite{SG:18} which offers optimized heuristics for the latter. As backend we use the state-of-the-art machine learning library \emph{PyTorch}~\cite{PGM+:19}, version 1.11.0, to do all operations between tensors and then perform backpropagation on them. We use \emph{pgmpy}~\citep{ankan2015pgmpy} for reading and moralizing BNs.  %For each MRF potential we use vectorized operations to obtain all metrics of interest at once, instead of looping over each parameter $\theta_i$.

All experiments were run on a 4-core i5-6600 3.3GHz Intel workstation with 16GB RAM.

\subsection{Risk Assessment for Humanitarian Crises and Disasters}

Similarly to \citet{qazi2021assessment}, we construct a BN model to assess the country-level risk
associated with  humanitarian crises and disasters. The data was collected from INFORM \citep{INFORM} and consists of 20 drivers of disaster risk covering natural, human, socio-economic, institutional and infrastructure factors that influence the country-level risk of a disaster, together with a final country risk index which summarizes how exposed a country is to the possibility of a humanitarian disaster. A full list of the variables can be found at \citet{INFORM}. All variables take values between zero and ten and have been discretized into three categories (low/0, medium/1, high/2) using the equal-length method. The dataset comprises 190 countries.

\begin{figure}
	\begin{center}
		\includegraphics[scale = 0.6]{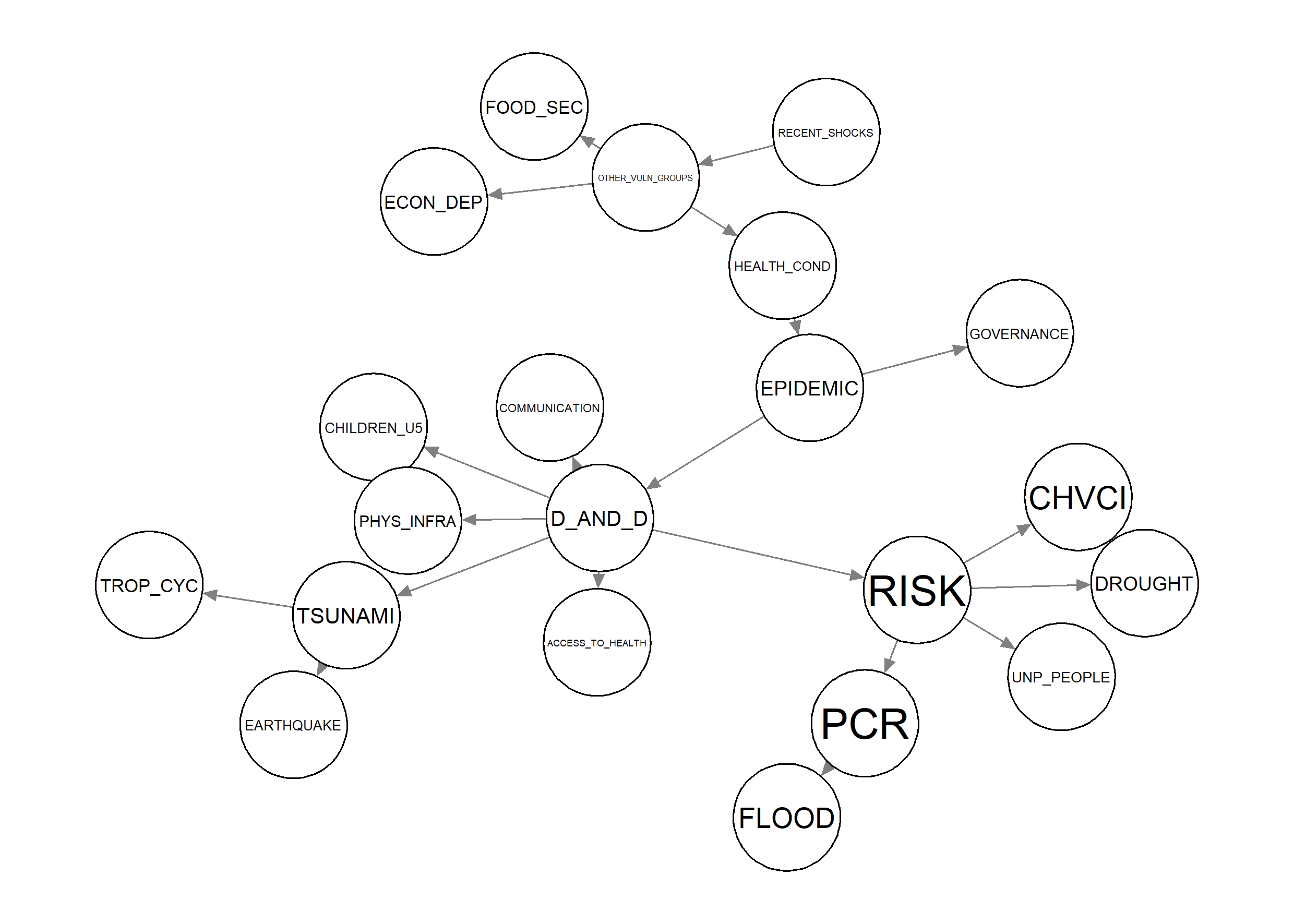}
	\end{center}
\caption{BN learned over the \citet{INFORM} dataset for country-level disaster risk.\label{fig:bn}}
\end{figure}

\begin{figure}
	\begin{center}
		\includegraphics[scale = 0.4]{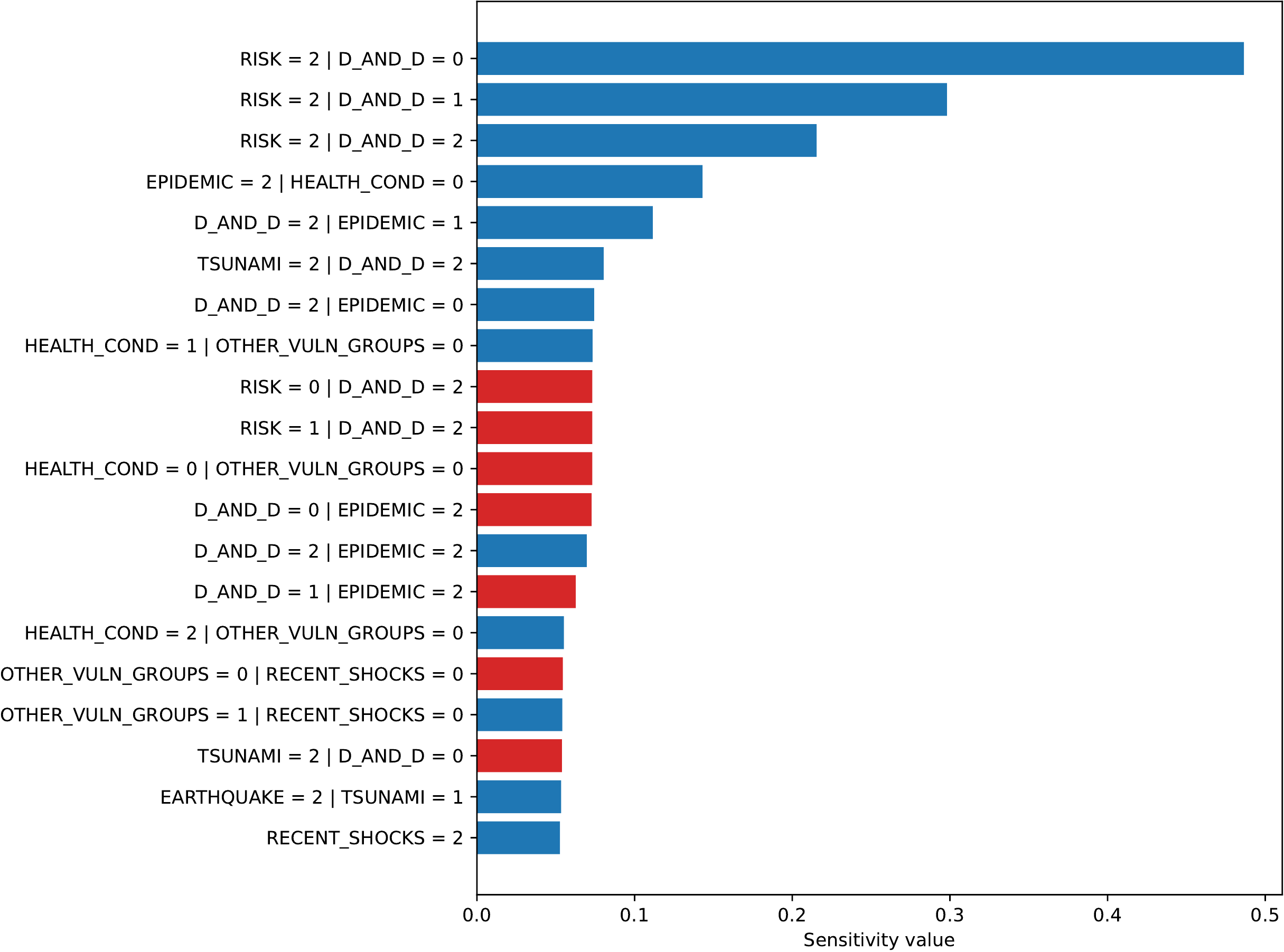}
	\end{center}
	\caption{Top 20 most influential parameters for the humanitarian crisis network, color-coded by the sign of $f'(\theta_i)$. The probability of interest is $P(\mbox{RISK} = \mbox{high} \;|\; \mbox{EARTHQUAKE} = \mbox{high})$.}
	\label{fig:many_parameters_plot}
\end{figure}

A BN is learned using the \texttt{hc} function of the \texttt{bnlearn} package and is reported in Figure \ref{fig:bn}. As an illustration of the YODO method, we compute here all sensitivity measures for the conditional probability of a high disaster risk (RISK = 2) conditional on a high risk of flooding (FLOOD = 2). Computing all metrics for all 183 network parameters with our method took only 0.055 seconds. The results  are reported in Table \ref{tab:many_parameter_study} for the 20 most influential parameters according to the sensitivity value. It can be noticed that the most influential parameters come from the conditional distributions of the overall risk given the development and deprivation index (D\_AND\_D), as well as from the conditional distribution of the flooding index given a projected conflict risk index (PCR) equal to low. As an additional illustration,  Figure \ref{fig:many_parameters_plot} reports the sensitivity value of the parameters for the output conditional probability of a high risk given a high risk of earthquake. The blue color is associated to positive values of the sensitivity value, the red color for negative ones. Out of 183 network parameters, 30 had a sensitivity value of zero, meaning that they had no effect on the probability of interest.

\begin{table} \tiny
	\begin{tabular}{llllll}
\toprule
{} &                  Value & Sens. value & Proximity & $2^{\mbox{nd}}$ deriv. &
Largest $1^{\mbox{st}}$ deriv. \\
Parameter                                 &                        &            
&           &                   &                          \\
\midrule
RISK = high $\vert$ D\_AND\_D = low                    &               $0.0012$ 
&           $0.914$ &   $0.056$ &          $-1.437$ &                  $0.916$ 
\\
FLOOD = high $\vert$ PCR = low                       &                $0.107$ & 
$0.722$ &  $0.0534$ &           $4.059$ &                  $1.475$ \\
FLOOD = medium $\vert$ PCR = low                       &                $0.469$ 
&           $0.645$ &   $0.718$ &           $3.238$ &                 $\infty$ 
\\
FLOOD = low $\vert$ PCR = low                       &                $0.425$ &  
$0.645$ &   $0.718$ &           $3.238$ &                 $\infty$ \\
RISK = high $\vert$ D\_AND\_D = high                    &                 $0.34$
&           $0.555$ &   $0.731$ &          $-0.387$ &                  $0.714$ 
\\
RISK = high $\vert$ D\_AND\_D = medium                    &               
$0.0686$ &           $0.467$ &   $1.002$ &          $-0.295$ &                  
$0.488$ \\
EPIDEMIC = high $\vert$ HEALTH\_COND = low            &                $0.148$ &
$0.295$ &   $1.167$ &          $-0.231$ &                  $0.332$ \\
D\_AND\_D = high $\vert$ EPIDEMIC = medium                &               
$0.0742$ &           $0.238$ &   $1.834$ &          $-0.133$ &                  
$0.249$ \\
PCR = high $\vert$ RISK = medium                        &                $0.278$
&           $0.226$ &     $0.6$ &           $0.395$ &                  $0.394$ 
\\
PCR = high $\vert$ RISK = low                        &               $0.0266$ & 
$0.204$ &   $0.694$ &           $0.322$ &                  $0.213$ \\
FLOOD = high $\vert$ PCR = high                       &                $0.509$ &
$0.196$ &   $0.475$ &           $-0.46$ &                  $1.211$ \\
FLOOD = high $\vert$ PCR = medium                       &                $0.136$
&           $0.167$ &   $0.787$ &            $0.25$ &                  $0.206$ 
\\
D\_AND\_D = high $\vert$ EPIDEMIC = high                &                $0.787$
&           $0.159$ &   $4.159$ &         $-0.0459$ &                  $0.202$ 
\\
D\_AND\_D = high $\vert$ EPIDEMIC = low                &               $0.0411$ 
&           $0.153$ &   $2.984$ &         $-0.0625$ &                  $0.156$ 
\\
RISK = low $\vert$ D\_AND\_D = high                    &               $0.0208$ 
&           $0.151$ &   $2.319$ &         $-0.0796$ &                  $0.274$ 
\\
HEALTH\_COND = medium $\vert$ OTHER\_VULN\_GROUPS = low   &                 
$0.05$ &           $0.151$ &   $3.026$ &          $-0.061$ &                  
$0.154$ \\
HEALTH\_COND = low $\vert$ OTHER\_VULN\_GROUPS = low   &                $0.949$ 
&            $0.15$ &   $3.036$ &         $-0.0606$ &                  $0.154$ 
\\
PCR = low $\vert$ RISK = high                        &              $0.00521$ & 
$0.15$ &   $3.023$ &         $-0.0609$ &                  $0.236$ \\
D\_AND\_D = medium $\vert$ EPIDEMIC = high                &                
$0.176$ &           $0.148$ &   $5.393$ &         $-0.0338$ &                   
$0.18$ \\
PCR = high $\vert$ RISK = high                        &                $0.943$ &
$0.148$ &   $3.092$ &         $-0.0588$ &                  $0.224$ \\
PCR = medium $\vert$ RISK = high                        &               $0.0521$
&           $0.146$ &   $3.144$ &         $-0.0573$ &                   $0.22$ 
\\
D\_AND\_D = low $\vert$ EPIDEMIC = high                &               $0.0368$ 
&           $0.144$ &   $2.859$ &         $-0.0626$ &                  $0.231$ 
\\
FLOOD = low $\vert$ PCR = medium                       &                $0.103$ 
&           $0.144$ &   $0.955$ &           $0.187$ &                  $0.165$ 
\\
FLOOD = medium $\vert$ PCR = medium                       &                
$0.762$ &           $0.144$ &   $0.955$ &           $0.187$ &                  
$0.565$ \\
RISK = medium $\vert$ D\_AND\_D = high                    &                 
$0.64$ &           $0.137$ &    $1.99$ &         $-0.0869$ &                  
$0.175$ \\
PCR = low $\vert$ RISK = medium                        &                $0.422$ 
&            $0.12$ &   $1.399$ &           $0.112$ &                  $0.226$ 
\\
HEALTH\_COND = high $\vert$ OTHER\_VULN\_GROUPS = low   &  $9.083 \cdot 10^{-4}$
&           $0.113$ &   $4.302$ &         $-0.0348$ &                  $0.113$ 
\\
OTHER\_VULN\_GROUPS = low $\vert$ RECENT\_SHOCKS = low &                 $0.82$ 
&           $0.112$ &   $4.418$ &         $-0.0337$ &                  $0.118$ 
\\
OTHER\_VULN\_GROUPS = medium $\vert$ RECENT\_SHOCKS = low &                
$0.173$ &           $0.111$ &   $4.454$ &         $-0.0333$ &                  
$0.117$ \\
PCR = low $\vert$ RISK = low                        &                $0.947$ &  
$0.109$ &   $1.578$ &          $0.0928$ &                  $0.115$ \\
\bottomrule
\end{tabular}

	\caption{Four sensitivity metrics for the top 30 parameters of the humanitarian crisis network (ranked by their sensitivity value), when the probability of interest is $P(\mbox{RISK} = \mbox{high} \;|\; \mbox{FLOOD} = \mbox{high})$. }
	\label{tab:many_parameter_study}
\end{table}

\subsection{Performance Study over Medium to Very Large Networks}

We further run our method over the 10 Bayesian networks considered in \citet{scutari2019learns}. As a baseline we use numerical estimation of each sensitivity value via finite differences, whereby we slightly perturb each parameter $\theta_i$ and measure the impact on $f$. As a probability of interest we set $P(A = a | B = b)$, where $A, B, a, b$ were two variables and two levels picked at random, respectively, and each timing is the average of three independent runs. Results are reported in Table~\ref{tab:many_network_study}, which shows that YODO outperforms the baseline by several orders of magnitude.

\begin{table} \small
	\begin{tabular}{lrrrrrr}
\toprule
{} &  \#nodes &  \#arcs &  \#parameters &  Treewidth &  Time (fin. diff.) &  
Time (ours) \\
Network    &          &         &               &            &                  
&              \\
\midrule
child      &       20 &      30 &           344 &          3 &           
2.188733 &     0.017727 \\
water      &       32 &     123 &         13484 &         10 &         
337.189158 &     0.054150 \\
alarm      &       37 &      65 &           752 &          4 &          
10.203079 &     0.034412 \\
hailfinder &       56 &      99 &          3741 &          4 &          
98.040870 &     0.053667 \\
hepar2     &       70 &     158 &          2139 &          6 &         
169.150984 &     0.093187 \\
win95pts   &       76 &     225 &          1148 &          8 &          
38.674632 &     0.113214 \\
pathfinder &      109 &     208 &         97851 &          6 &        
8596.810546 &     0.188448 \\
munin1     &      186 &     354 &         19226 &         11 &      
113928.398017 &    14.394249 \\
andes      &      223 &     626 &          2314 &         17 &         
252.286060 &     0.299587 \\
pigs       &      441 &     806 &          8427 &         10 &        
3213.308629 &     0.521486 \\
\bottomrule
\end{tabular}

	\caption{Our method was applied to 10 Bayesian networks, here sorted by number of nodes. All times are in seconds. The times for the baseline (second-to-last column) were estimated as the total number of parameters in the network times the time needed to numerically estimate one sensitivity value. Treewidths were found with the \emph{NetworkX} graph library~\cite{HSS:08}.}
	\label{tab:many_network_study}
\end{table}

\section{Discussion}
\label{sec:discussion}

We demonstrated the use of automatic differentiation in the area of BNs and more specifically in the study of how sensitive they are to variations of their parameters. The novel algorithms are freely available in Python and are planned to be included in the next release of the \texttt{bnmonitor} R package \citep{leonelli2021sensitivity}. Their efficiency was demonstrated through a simulation study and their use in practice was illustrated through a BN in the field of risk assesment for humanitarian crises.

Although YODO is specifically designed to compute the coefficients of the sensitivity function in Equation \ref{eq:sens}, it further addresses two additional problems in sensitivity analysis. First, it is able to quickly find which parameters do have an effect on the output probability of interest, which is usually called the \textit{parameter sensitivity set} \citep{coupe2002properties}. Second, we identify whether a parameter change leads to a monotonically increasing or decreasing sensitivity function, as already addressed in \citet{bolt2017structure}. Although the above-cited works only require the structure of the network, YODO yields an efficient way to tackle the same problems.

\subsection*{Future Work}

Because of the space constraint we only focused on one-way sensitivity analysis but, because of their efficiency, the proposed methods could be generalized to multi-way sensitivity analysis where more than one parameter is varied contemporaneously. \citet{bolt2014local} introduced the \textit{maximum/minimum n-way sensitivity value} which bounds the effect of n-way variations of parameters and demonstrated that it can be easily derived from the sensitivity values of one-way sensitivity analyses. Therefore, our methods could be extended to also efficiently compute the joint effect of variations of parameters, known to be computationally challenging~\citep{uai2004,uai2000}.

Another possible extension of the algorithms introduced here would be to compute the so-called \textit{admissible deviation} \citep{uai2001}. This consists of finding a pair of numbers $(\alpha,\beta)$ that describe the shifts to smaller values and to larger values, respectively, that are allowed in the parameter under study without inducing a change in the most likely value of the output variable. For a parameter with an original value of $\theta_i^0$, the admissible deviation $(\alpha,\beta)$ thus indicates that the parameter can be safely varied within the interval $(\theta_i^0-\alpha,\theta_i^0+\beta)$. These values can be straightforwardly found by identifying the intersections of the sensitivity functions associated to different values of the output variable.

\bibliography{references}

\begin{thebibliography}{33}
\providecommand{\natexlab}[1]{#1}
\providecommand{\url}[1]{\texttt{#1}}
\expandafter\ifx\csname urlstyle\endcsname\relax
  \providecommand{\doi}[1]{doi: #1}\else
  \providecommand{\doi}{doi: \begingroup \urlstyle{rm}\Url}\fi

\bibitem[Ankan and Panda(2015)]{ankan2015pgmpy}
A.~Ankan and A.~Panda.
\newblock pgmpy: Probabilistic graphical models using python.
\newblock In \emph{Proceedings of the 14th Python in Science Conference (SCIPY
  2015)}. Citeseer, 2015.

\bibitem[Ballester-Ripoll and Leonelli(2022)]{ballester2022computing}
R.~Ballester-Ripoll and M.~Leonelli.
\newblock Computing {S}obol indices in probabilistic graphical models.
\newblock \emph{Reliability Engineering \& System Safety}, 225:\penalty0
  108573, 2022.

\bibitem[Bolt and Renooij(2014)]{bolt2014local}
J.~H. Bolt and S.~Renooij.
\newblock Local sensitivity of {B}ayesian networks to multiple simultaneous
  parameter shifts.
\newblock In \emph{European Workshop on Probabilistic Graphical Models}, pages
  65--80, 2014.

\bibitem[Bolt and Renooij(2017)]{bolt2017structure}
J.~H. Bolt and S.~Renooij.
\newblock Structure-based categorisation of {B}ayesian network parameters.
\newblock In \emph{European Conference on Symbolic and Quantitative Approaches
  to Reasoning and Uncertainty}, pages 83--92, 2017.

\bibitem[Castillo et~al.(1997)Castillo, Guti{\'e}rrez, and
  Hadi]{castillo1997sensitivity}
E.~Castillo, J.~M. Guti{\'e}rrez, and A.~S. Hadi.
\newblock Sensitivity analysis in discrete {B}ayesian networks.
\newblock \emph{IEEE Transactions on Systems, Man, and Cybernetics-Part A:
  Systems and Humans}, 27\penalty0 (4):\penalty0 412--423, 1997.

\bibitem[Chan and Darwiche(2004)]{uai2004}
H.~Chan and A.~Darwiche.
\newblock Sensitivity analysis in {B}ayesian networks: From single to multiple
  parameters.
\newblock In \emph{Proceedings of the 20th UAI Conference}, pages 67--75, 2004.

\bibitem[Chan and Darwiche(2005)]{chan2005distance}
H.~Chan and A.~Darwiche.
\newblock A distance measure for bounding probabilistic belief change.
\newblock \emph{International Journal of Approximate Reasoning}, 38\penalty0
  (2):\penalty0 149--174, 2005.

\bibitem[Coup{\'e} and Van~der Gaag(2002)]{coupe2002properties}
V.~M. Coup{\'e} and L.~C. Van~der Gaag.
\newblock Properties of sensitivity analysis of {B}ayesian belief networks.
\newblock \emph{Annals of Mathematics and Artificial Intelligence}, 36\penalty0
  (4):\penalty0 323--356, 2002.

\bibitem[Coup{\'e} et~al.(2000)Coup{\'e}, Van Der~Gaag, and
  Habbema]{coupe2000sensitivity}
V.~M. Coup{\'e}, L.~C. Van Der~Gaag, and J.~D.~F. Habbema.
\newblock Sensitivity analysis: an aid for belief-network quantification.
\newblock \emph{The Knowledge Engineering Review}, 15\penalty0 (3):\penalty0
  215--232, 2000.

\bibitem[Darwiche(2003)]{darwiche2003differential}
A.~Darwiche.
\newblock A differential approach to inference in {B}ayesian networks.
\newblock \emph{Journal of the ACM (JACM)}, 50\penalty0 (3):\penalty0 280--305,
  2003.

\bibitem[Darwiche(2009)]{Darwiche2009}
A.~Darwiche.
\newblock \emph{Modeling and reasoning with {B}ayesian networks}.
\newblock Cambridge University Press, 2009.

\bibitem[Douglas-Smith et~al.(2020)Douglas-Smith, Iwanaga, Croke, and
  Jakeman]{Douglas2020}
D.~Douglas-Smith, T.~Iwanaga, B.~F. Croke, and A.~J. Jakeman.
\newblock Certain trends in uncertainty and sensitivity analysis: An overview
  of software tools and techniques.
\newblock \emph{Environmental Modelling \& Software}, 124:\penalty0 104588,
  2020.

\bibitem[Goerlandt and Islam(2021)]{goerlandt2021bayesian}
F.~Goerlandt and S.~Islam.
\newblock {A Bayesian Network risk model for estimating coastal maritime
  transportation delays following an earthquake in British Columbia}.
\newblock \emph{Reliability Engineering \& System Safety}, 214:\penalty0
  107708, 2021.

\bibitem[Hagberg et~al.(2008)Hagberg, Schult, and Swart]{HSS:08}
A.~A. Hagberg, D.~A. Schult, and P.~J. Swart.
\newblock Exploring network structure, dynamics, and function using {NetworkX}.
\newblock In \emph{Proceedings of the 7th Python in Science Conference}, pages
  11--15, 2008.

\bibitem[{INFORM}(2022)]{INFORM}
{INFORM}.
\newblock Index for risk management.
\newblock Retrieved from https://drmkc.jrc.ec.europa.eu/inform-index, 2022.

\bibitem[Kjaerulff and van~der Gaag(2000)]{uai2000}
U.~Kjaerulff and L.~van~der Gaag.
\newblock Making sensitivity analysis computationally efficient.
\newblock In \emph{Proceedings of the 17th UAI Conference}, pages 317--325,
  2000.

\bibitem[Kwisthout and van~der Gaag(2008)]{uai2008}
J.~Kwisthout and L.~van~der Gaag.
\newblock The computational complexity of sensitivity analysis and parameter
  tuning.
\newblock In \emph{Proceedings of the 24th UAI Conference}, pages 349--356,
  2008.

\bibitem[Laskey(1995)]{laskey1995sensitivity}
K.~B. Laskey.
\newblock Sensitivity analysis for probability assessments in {B}ayesian
  networks.
\newblock \emph{IEEE Transactions on Systems, Man, and Cybernetics},
  25\penalty0 (6):\penalty0 901--909, 1995.

\bibitem[Leonelli and Riccomagno(2018)]{leonelli2018geometric}
M.~Leonelli and E.~Riccomagno.
\newblock A geometric characterisation of sensitivity analysis in monomial
  models.
\newblock \emph{arXiv:1901.02058}, 2018.

\bibitem[Leonelli et~al.(2017)Leonelli, G{\"o}rgen, and
  Smith]{leonelli2017sensitivity}
M.~Leonelli, C.~G{\"o}rgen, and J.~Q. Smith.
\newblock Sensitivity analysis in multilinear probabilistic models.
\newblock \emph{Information Sciences}, 411:\penalty0 84--97, 2017.

\bibitem[Leonelli et~al.(2021)Leonelli, Ramanathan, and
  Wilkerson]{leonelli2021sensitivity}
M.~Leonelli, R.~Ramanathan, and R.~L. Wilkerson.
\newblock Sensitivity and robustness analysis in {B}ayesian networks with the
  bnmonitor {R} package.
\newblock \emph{arXiv:2107.11785}, 2021.

\bibitem[Li and Mahadevan(2018)]{li2018sensitivity}
C.~Li and S.~Mahadevan.
\newblock Sensitivity analysis of a {B}ayesian network.
\newblock \emph{ASCE-ASME J Risk and Uncert in Engrg Sys Part B Mech Engrg},
  4\penalty0 (1), 2018.

\bibitem[Makaba et~al.(2021)Makaba, Doorsamy, and Paul]{makaba2021bayesian}
T.~Makaba, W.~Doorsamy, and B.~S. Paul.
\newblock Bayesian network-based framework for cost-implication assessment of
  road traffic collisions.
\newblock \emph{International journal of intelligent transportation systems
  research}, 19\penalty0 (1):\penalty0 240--253, 2021.

\bibitem[Paszke et~al.(2019)Paszke, Gross, Massa, and et~al.]{PGM+:19}
A.~Paszke, S.~Gross, F.~Massa, and et~al.
\newblock {P}y{T}orch: An imperative style, high-performance deep learning
  library.
\newblock In H.~Wallach, H.~Larochelle, A.~Beygelzimer, F.~d\textquotesingle
  Alch\'{e}-Buc, E.~Fox, and R.~Garnett, editors, \emph{Advances in Neural
  Information Processing Systems}, pages 8024--8035. Curran Associates, Inc.,
  2019.

\bibitem[Qazi and Simsekler(2021)]{qazi2021assessment}
A.~Qazi and M.~C.~E. Simsekler.
\newblock Assessment of humanitarian crises and disaster risk exposure using
  data-driven {B}ayesian networks.
\newblock \emph{International Journal of Disaster Risk Reduction}, 52:\penalty0
  101938, 2021.

\bibitem[Renooij(2014)]{renooij2014co}
S.~Renooij.
\newblock Co-variation for sensitivity analysis in {B}ayesian networks:
  properties, consequences and alternatives.
\newblock \emph{International Journal of Approximate Reasoning}, 55\penalty0
  (4):\penalty0 1022--1042, 2014.

\bibitem[Robeva and Seigal(2018)]{RS:18}
E.~Robeva and A.~Seigal.
\newblock {Duality of graphical models and tensor networks}.
\newblock \emph{Information and Inference: A Journal of the IMA}, 8\penalty0
  (2):\penalty0 273--288, 06 2018.

\bibitem[Rohmer(2020)]{rohmer2020uncertainties}
J.~Rohmer.
\newblock Uncertainties in conditional probability tables of discrete
  {B}ayesian belief networks: A comprehensive review.
\newblock \emph{Engineering Applications of Artificial Intelligence},
  88:\penalty0 103384, 2020.

\bibitem[Scutari et~al.(2019)Scutari, Graafland, and
  Guti{\'e}rrez]{scutari2019learns}
M.~Scutari, C.~E. Graafland, and J.~M. Guti{\'e}rrez.
\newblock Who learns better {B}ayesian network structures: Accuracy and speed
  of structure learning algorithms.
\newblock \emph{International Journal of Approximate Reasoning}, 115:\penalty0
  235--253, 2019.

\bibitem[Smith and Gray(2018)]{SG:18}
D.~G.~A. Smith and J.~Gray.
\newblock opt\_einsum - a {P}ython package for optimizing contraction order for
  einsum-like expressions.
\newblock \emph{Journal of Open Source Software}, 3\penalty0 (26):\penalty0
  753, 2018.

\bibitem[van~der Gaag and Renooij(2001)]{uai2001}
L.~van~der Gaag and S.~Renooij.
\newblock Analysing sensitivity data from probabilistic networks.
\newblock In \emph{Proceedings of the 18th UAI Conference}, pages 530--537,
  2001.

\bibitem[Van Der~Gaag et~al.(2007)Van Der~Gaag, Renooij, and
  Coup{\'e}]{van2007sensitivity}
L.~C. Van Der~Gaag, S.~Renooij, and V.~M. Coup{\'e}.
\newblock Sensitivity analysis of probabilistic networks.
\newblock In \emph{Advances in probabilistic graphical models}, pages 103--124.
  Springer, 2007.

\bibitem[Zio et~al.(2022)Zio, Mustafayeva, and Montanaro]{zio2022bayesian}
E.~Zio, M.~Mustafayeva, and A.~Montanaro.
\newblock A {B}ayesian belief network model for the risk assessment and
  management of premature screen-out during hydraulic fracturing.
\newblock \emph{Reliability Engineering \& System Safety}, 218:\penalty0
  108094, 2022.

\end{thebibliography}
\end{document}